\documentclass{article}
\usepackage[a4paper, left=1.7cm, right=1.7cm, top=1in, bottom=1in]{geometry}
\usepackage[T1]{fontenc}
\usepackage[utf8]{inputenc}
\usepackage[german,english]{babel}

\usepackage{graphicx}

\usepackage{authblk}

\newcommand{\rev}[1]{#1}
\newcommand{\revTransNote}[1]{{\small{\fontfamily{phv}\selectfont{#1}}}}

\usepackage{amsmath}
\usepackage{amssymb}
\usepackage[super]{nth}
\usepackage{paracol}
\newenvironment{german}
  {\begin{rightcolumn}\begin{otherlanguage}{german}}
  {\end{otherlanguage}\end{rightcolumn}}
\newenvironment{english}
  {\begin{leftcolumn*}\begin{otherlanguage}{english}}
  {\end{otherlanguage}\end{leftcolumn*}}
\newenvironment{translated}
  {\begin{paracol}{2}}
  {\end{paracol}}
  
\global\long\def\mH{\mathtt{H}}

\usepackage{url}
\usepackage{hyperref}
\hypersetup{pdfinfo={
Title={Translation of Zur Ermittlung eines Objektes aus zwei Perspektiven mit innerer Orientierung by Erwin Kruppa (1913)},
Author={Guillermo Gallego, Elias Mueggler and Peter Sturm},
Subject={Photogrammetry, Computer Vision, Robotics},
Keywords={Camera calibration, epipolar geometry, 5-point algorithm, projective geometry, Kruppa's equations, absolute conic}
}}

\usepackage{natbib}
\usepackage[final]{pdfpages}

\title{Translation of ``Zur Ermittlung eines Objektes aus zwei Perspektiven mit innerer Orientierung'' by Erwin Kruppa (1913)\footnote{We thank the publishing house of the Austrian Academy of Sciences (ÖAW, \url{https://verlag.oeaw.ac.at/}) for granting permission to reproduce the original paper by Erwin Kruppa.}}
\author[1]{Guillermo Gallego}
\author[1]{Elias Mueggler}
\author[2]{Peter Sturm}
\affil[1]{University of Zurich, Zurich, Switzerland.}
\affil[2]{Inria and Univ. Grenoble Alpes, Lab. LJK, Grenoble, France.}
\date{}

\begin{document}

\maketitle

\begin{abstract}
Erwin Kruppa's 1913 paper,
\begin{quote}
Erwin Kruppa, ``Zur Ermittlung eines Objektes aus zwei Perspektiven mit innerer Orientierung'', 
Sitzungsberichte der Mathematisch-Naturwissenschaftlichen Kaiserlichen Akademie der Wissenschaften, Vol.~122 (1913), pp. 1939--1948, 
\end{quote}
which may be translated as ``To determine a 3D object from two perspective views with known inner orientation'',
is a landmark paper in Computer Vision because it provides the first five-point algorithm for relative pose estimation.
%i.e., he showed that from five point matches between two calibrated images of a rigid object, the relative pose between the images can be computed up to a finite number of solutions.
%He showed that the relative pose between two calibrated images of a rigid object can be computed, up to a finite number of solutions, from five point matches between the images.
Kruppa showed that (a finite number of solutions for) the relative pose between two calibrated images of a rigid object can be computed from five point matches between the images.
Kruppa's work also gained attention in the topic of camera self-calibration, as presented in~\citep{Maybank92ijcv}.
Since the paper is still relevant today (more than a hundred citations within the last ten years) and the paper is not available online, we ordered a copy from the German National Library in Frankfurt and provide an English translation along with the German original.
We also adapt the terminology to a modern jargon 
and provide some clarifications (\revTransNote{highlighted in sans-serif font}).
For a historical review of geometric computer vision, the reader is referred to the recent survey paper~\citep{Sturm11caip}.
\end{abstract}

\section*{Summary of Kruppa's Approach}

It is well known (and was already known by the time Kruppa's paper was written) that the epipolar geometry between two \emph{uncalibrated} perspective images can be determined from seven point correspondences, up to three possible solutions in general. 
Kruppa showed that, when the images are \emph{calibrated}, five correspondences are sufficient. The epipolar geometry, combined with the calibration information, then allows to recover the relative pose of the two images (up to scale), in turn enabling to determine the shape of the 3D scene (also up to scale).
He proved that, in general, there are up to 11 solutions (22 possible relative poses, coming in 11 pairs of solutions corresponding to relative poses related by rotating one camera by 180 degrees about the baseline joining the cameras' optical centers, since such a rotation leaves the reconstructed 3D shape invariant). 
In 1988, it was shown by~\citep{Demazure88inria} that there are actually at most 10 distinct solutions. 
Later, \citep{Nister04pami} developed the first algorithm that guarantees to indeed provide at most 10 solutions.

Kruppa's approach can be briefly outlined as follows. 
As mentioned, it was already known how to compute the epipolar geometry from seven point correspondences.
When using only five such correspondences, two additional constraints are required. 
Kruppa's insight was that the knowledge of the cameras' calibration delivers such constraints.
In particular, knowing the calibration of a camera is equivalent to knowing the image of the absolute conic (IAC) in that camera. The absolute conic is invariant to similarity transformations and thus the IAC is invariant to camera displacements. 
Hence, the IAC is directly determined by the camera's intrinsic parameters and indeed, the IAC fully encodes the intrinsic parameters of a perspective camera.
Kruppa essentially formulated what could be called epipolar constraints for images of conics.
For images of points, the epipolar constraint states that the two image points lie on corresponding epipolar lines. 
As for images of conics, we have: each of the two image conics has two epipolar lines tangent to it -- these two pairs of epipolar lines must correspond to one another.
This observation gives rise to two constraints which, together with five point correspondences, allow to compute the epipolar geometry and, subsequently, the relative pose and the 3D shape of the scene (all these up to the stated number of possible solutions).

The term ``Kruppa's equations'' was probably coined by~\citep{Maybank92ijcv}, who
showed how to use the epipolar constraint for images of conics, for self-calibration: 
whereas Kruppa used the known calibration to compute epipolar geometry, Maybank and Faugeras showed that
known epipolar geometry allows to compute the camera calibration, even in the absence of a
calibration grid in the scene (to be precise, depending on the number of intrinsic parameters
to be computed, one needs to provide the epipolar geometry for more than one image pair).

\begin{translated}

\begin{english}
\section*{English Translation}
S. Finsterwalder proves in his paper ``\href{https://eudml.org/doc/144606}{The Geometric Foundations of Photogrammetry}'' 
(Annual report of the German Association of Mathematicians (1897), Vol. 6, pp. 1--42) %ISSN: 0012-0456; 1869-7135.
this theorem (p.~15):
\end{english}
\begin{german}
\section*{German Original}
S. Finsterwalder beweist in seinem Referate ``Die geometrischen Grundlagen der Photogrammetrie'' (Jahresbericht der Deutschen Mathematiker-Vereinigung~6 Bd., 2.~Heft) den Satz (p.~15):
\end{german}

\begin{english}
Two perspective views with known inner orientation \revTransNote{(calibration)} are sufficient to determine the object up to scale.
\end{english}
\begin{german}
Zwei Perspektiven mit bekannter innerer Orientierung reichen aus, um das Objekt bis auf den Maßstab zu bestimmen.
\end{german}

\begin{english}
The term ``inner orientation'' refers to the knowledge of the position of the center of projection relative to the perspective view.
\end{english}
\begin{german}
Dabei versteht man unter ``innerer Orientierung'' die Kenntnis der relativen Lage des Projektionszentrums zur Perspektive.
\end{german}

\begin{english}
To compute the orientation \revTransNote{(relative pose)} of the perspective views, 
Mr. Finsterwalder determines the Hauckian kernel points \revTransNote{(epipoles)} from 7 pairs of corresponding image points.
However, a simple count of constants \revTransNote{(degrees-of-freedom)}\footnote{This remark was made by Mr. Finsterwalder in his work: ``\href{http://biodiversitylibrary.org/page/35671676}{A basic problem in Photogrammetry}'' 
(Treatises of the Mathematical-Physics Class of the Royal Bavarian Academy of Sciences, vol XXII., Issue~II, p.~230.).
cf. \emph{E. Kruppa}: ``On some Orientation Problems in Photogrammetry'' (These proceedings, vol.~CXXI, Issue.~IIa, p.~9, Remark).
}
reveals that the epipoles are determined from 5 pairs of corresponding image points 
if the intrinsic calibration is assumed to be known.
This work is devoted to this fact.
It yields the following theorems:
\end{english}
\begin{german}
Um die Orientierung der Perspektiven durchzuführen, bestimmt Herr Finsterwalder die Hauck'schen Kernpunkte aus 7~Bildpaaren von Objektpunkten.
Nun sind aber, wie aus einer einfachen Konstantenzählung\footnote{Diese Bemerkung macht Herr Finsterwalder in seiner Arbeit: 
Eine Grundaufgabe der Photogrammetrie (Abhandlungen der Mathematisch-Physikalischen Klasse der Königlich Bayerischen Akademie der Wissenschaften 
II.~Kl., XXII.~Bd., II.~Abt., p.~230). Vergl. \emph{E. Kruppa}: Über einige Orientierungsprobleme der Photogrammetrie (Diese Berichte, Bd.~CXXI, Abt.~IIa, p.~9, Anmerkung).} 
folgt, die Kernpunkte bereits durch 5~Bildpaare von Objektpunkten bestimmt, wenn man noch die innere Orientierung heranzieht.
Die vorliegende Arbeit ist dieser Tatsache gewidmet. 
Es ergeben sich folgende Sätze:
\end{german}

\begin{english}
\textbf{\nth{1} Theorem:} A three-dimensional pentagon can be determined, up to scale, by two arbitrary perspective views with known inner orientation [in general, there are 22~solutions, possibly complex].
\end{english}
\begin{german}
1. Satz: Ein räumliches Fünfeck ist durch die willkürliche Annahme von 2~Perspektiven mit innerer Orientierung, abgsehen vom Maßstab, bestimmbar [i. a. auf 22~Arten, ohne Rücksicht auf die Realität].
\end{german}

\begin{english}
\textbf{\nth{2} Theorem:} A three-dimensional heptagon can be determined, up to scale, by two arbitrary perspective views given the inner orientation of one and the distance to the center of projection of the other [in general, there are 24~solutions, possibly complex; see footnote~\ref{foot:thm2} on p.~\pageref{foot:thm2}].
\end{english}
\begin{german}
2. Satz: Ein räumliches Siebeneck ist durch die willkürliche Annahme von 2~Perspektiven, der inneren Orientierung der einen und der Distanz des Projektionszentrums der anderen, abgesehen vom Maßstab, bestimmbar [i. a. auf 24~Arten, ohne Rücksicht auf die Realität; man beachte die Anmerkung~\ref{foot:thm2} auf p.~\pageref{foot:thm2}].
\end{german}

%%%%%%%%%%%%%%%%%%%%%%%%%%%%%%%%%%%%%%%%%%%%%%%%%%%%%%%%%%%%%%%%%%%%%%555

\begin{english}
\section*{Proof of \nth{1} Theorem}
We denote the centers of the views with $o_1$ and $o_2$, and the corresponding image planes with $\epsilon$ and $\epsilon'$.
The line $\overline{o_1 o_2}$ intersects $\epsilon$ and $\epsilon'$ in $o$ and $o'$, respectively.
Points $o$ and $o'$ are known as the epipoles of the perspective views.
For every point $p^r$ that does not lie on the line $\overline{o_1 o_2}$,
its viewing rays $\overline{o_1 p^r}$ and $\overline{o_2 p^r}$ lie on a certain plane $\alpha$ \revTransNote{(the epipolar plane through $p^r$)}, 
and its projections $p$ and $p'$ lie, therefore, on the lines given by the intersections $\epsilon \cap \alpha$ and $\epsilon' \cap \alpha$ \revTransNote{(epipolar lines)}. 
The set of these intersection lines form the two ``kernel line pencils'' \revTransNote{(pencils of epipolar lines)} $o$ and $o'$. 
Since these pencils are projectively related to the pencil of (epipolar) planes $\alpha$, the following known theorem holds:
\end{english}
\begin{german}
\section*{Zu 1.}
Wir bezeichnen die Zentren der Perspektiven mit $o_1$ und $o_2$, die zugehörigen Bildebenen mit $\epsilon$ und $\epsilon'$.
$\overline{o_1 o_2}$ schneide $\epsilon$ in $o$, $\epsilon'$ in $o'$. 
$o, o'$ heissen bekanntlich die Kernpunkte der Perspektiven.
Für jeden Raumpunkt $p^r$, der nicht auf $\overline{o_1 o_2}$ liegt, liegen die Sehstrahlen $\overline{o_1 p^r}$ und $\overline{o_2 p^r}$ in einer bestimmten Ebene~$\alpha$, seine Bilder $p$ und $p'$ daher in den Schnittlinien $[\epsilon~\alpha]$ und $[\epsilon'~\alpha]$.
Die Gesamtheit dieser Schnittlinien bildet die beiden ``Kernstrahlbüschel'' $o$ und $o'$.
Da sie zum Büschel der Ebene~$\alpha$ perspektiv liegen, gilt der bekannte Satz:
\end{german}

\begin{english}
$a)$ The pairs of image points corresponding to object points 
are projected from the epipoles by means of projective pencils
\revTransNote{(i.e., the set of lines formed by the epipole and image points in one image 
is in projective relation to the analogous set of lines in the other image)}.
\end{english}
\begin{german}
$a)$ Die Bildpaare der Objektpunkte werden aus den Kernpunkten durch projektive Büschel projiziert.
\end{german}

\begin{english}
Our investigation is now based on the following addition:
\end{english}
\begin{german}
Unsere Untersuchung beruht nun auf folgendem Zusatz:
\end{german}

\begin{english}
\label{thm:EpipolarLineHomography}
$b)$ In the projectivity of both epipolar line pencils \revTransNote{(given by the epipolar line homography)}, 
the tangents that can be drawn from the epipoles to the images of a curve are corresponding lines.
\end{english}
\begin{german}
$b)$ In der Projektivität der beiden Kernstrahlbüschel entsprechen i. a. einander auch die Tangenten, die man aus den Kernpunkten an die Bilder einer Kurve legen kann.
\end{german}

\begin{english}
In fact, if a tangential plane $\alpha$ to the space curve $I^r$, with contact point $p^r$,
passes through $\overline{o_1 o_2}$, 
then the epipolar lines %``nuclear rays'' 
to $p$ and $p'$ are tangent to $I$ and $I'$, respectively.
\end{english}
\begin{german}
In der Tat, geht durch $\overline{o_1 o_2}$ eine Tangentialebene~$\alpha$ an die Raumkurve $I^r$ mit dem Berührungspunkt $p^r$, so sind die Kernstrahlen nach $p$ und $p'$ i. a. Tangenten an $I$ beziehungsweise $I'$.
\end{german}

\begin{english}
Let us apply the above statement to the Absolute Conic $I^r$.
Since the calibration of the views is given, 
the projections $I$ and $I'$ \revTransNote{(Images of the Absolute Conic (IACs))} are known.
In addition, for the determination of the object, 
we are given 5 pairs of image points $a\,b\,c\,d\,e \to a'\,b'\,c'\,d'\,e'$.
\end{english}
\begin{german}
Diesen Zusatz wenden wir auf den absoluten Kegelschnitt $I^r$ an.
Durch die innere Orientierung sind seine Bilder $I$ und $I'$ bekannt.
Außerdem stehen zur Ermittlung des Objekts 5~Bildpunktpaare $a\,b\,c\,d\,e \to a'\,b'\,c'\,d'\,e'$ zur Verfügung.
\end{german}

\begin{english}
According to the above statement, the determination of the epipoles 
consists of finding a point $o$ in $\epsilon$ and a point $o'$ in $\epsilon'$ such that the projectivity \revTransNote{(homography)}
\begin{equation*}
o(a\,b\,c\,d\,e\,I) \barwedge o'(a'\,b'\,c'\,d'\,e'\,I') 
\end{equation*}
holds.
\revTransNote{(The 5 lines $\overline{oa}\mapsto \overline{o'a'}$, 
$\;\overline{ob}\mapsto \overline{o'b'}$, etc. 
as well as the epipolar lines tangent to $I$ and $I'$ are related by a projective transformation, i.e., a homography. 
See Fig.~\ref{fig:epipolardiagram}).}
\end{english}
\begin{german}
Nach dem Voranstehenden kommt die Bestimmung der Kernpunkte darauf hinaus, in $\epsilon$ einen Punkt~$o$ und in $\epsilon'$ einen Punkt~$o'$ zu finden, so daß die Projektivität
\begin{equation*}
o(a\,b\,c\,d\,e\,I) \barwedge o'(a'\,b'\,c'\,d'\,e'\,I') 
\end{equation*}
besteht.
\end{german}

\begin{figure*}[t]
\centering
\includegraphics[width=0.8\linewidth]{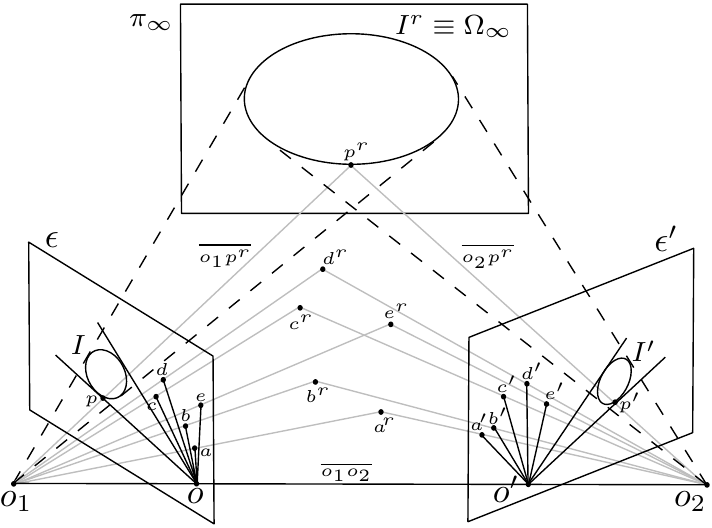}
\caption{\revTransNote{Diagram with the geometric configuration considered in Theorem 1. 
The theorem is based on the projective correspondence between the pencils of epipolar lines through the epipoles $o$ and $o'$.
Using modern notation, the absolute conic is $\Omega_\infty$ and the plane at infinity is $\pi_\infty$.}}
\label{fig:epipolardiagram}
\end{figure*}

\begin{english}
This is a generalization of the known ``Projectivity problem''\footnote{Cf. R. Sturm: `\href{http://gdz.sub.uni-goettingen.de/dms/load/img/?PID=GDZPPN002240270}{The projectivity problem and its application to 2nd degree surfaces} (\href{http://www.springerlink.com/content/1432-1807/}{Mathematische Annalen}, Vol. 1, pp. 533--574 (1869)).\par
Hesse: ``\href{https://eudml.org/doc/147890}{The cubic equation on which the solution to the problem of the homography of M. Chasles depends}''.
(Journal of pure and applied mathematics, Vol.~62, p. 188--192 (1863))\par
H. v. Sanden: ``\href{https://gso.gbv.de/DB=2.1/PPNSET?PPN=311071058}{Determination of the epipoles in Photogrammetry}''. (University of Göttingen Dissertation 1908)} for 7 point pairs, 
in that here, irreducible second-order curves replace two of the point pairs.
\end{english}
\begin{german}
Es handelt sich also um eine Verallgemeinerung des bekannten ``Problems der Projektivität''\footnote{Vgl. R. Sturm: Das Problem der Projektivität und seine Anwendung auf Flächen 2.~Gr. (Math. Ann.~I, 1869).\par
Hesse: 
Die cubische Gleichung, von welcher die Lösung des Problems der Homographie von M. Chasles abhängt. 
(Journal für die reine und angewandte Mathematik~62, p.~188)\par
H. v. Sanden: Bestimmung der Kernpunkte in der Photogrammetrie. (Göttinger Dissertation 1908)} für 7~Punktpaare, indem hier an Stelle von 2~Punktpaaren irreduzible Kurven 2.~Klasse treten. 
\end{german}

\begin{english}
Let us use analytical geometry.
Let $a\,b\,c$ and $a'\,b'\,c'$ be the fundamental triangles, each of a system of projective coordinates in $\epsilon$ and $\epsilon'$; 
it is $a(1,0,0)$; $b(0,1,0)$; $c(0,0,1)$; $d(d_1,d_2,d_3)$; $e(e_1,e_2,e_3)$; 
$I: \sum_{ik} a_{ik} y_i y_k = 0$; $a_{ik} = a_{ki}$.
And similarly for $\epsilon'$.
The coordinates of the desired epipoles are $o(x_1\,x_2\,x_3)$ and $o'(x'_1\,x'_2\,x'_3)$.
\end{english}
\begin{german}
Wir bedienen uns der analytischen Geometrie.
$a\,b\,c$ und $a'\,b'\,c'$ seien die Fundamentaldreiecke je eines \rev{Sys-tems} projektiver Koordinaten in $\epsilon$ und $\epsilon'$; es sei $a(1,0,0)$; $b(0,1,0)$; $c(0,0,1)$; d$(d_1,d_2,d_3)$; $e(e_1,e_2,e_3)$; 
$I: \sum a_{ik} y_i y_k = 0$; $a_{ik} = a_{ki}$.
Entsprechendes gelte in $\epsilon'$.
Die Koordinaten der gesuchten Kernpunkte seien $o(x_1\,x_2\,x_3)$ und $o'(x'_1\,x'_2\,x'_3)$.
\end{german}

\begin{english}
Let us now establish the condition that the mentioned tuples of 7 rays are projectively related.
We establish the equivalent condition that the intersections of these pairs of rays with $\overline{a\,b}$ and $\overline{a'\,b'}$, respectively, are projectively related.
The coordinates of these seven intersection points are, in order:\footnote{Notation: $(d\,x)_1=d_2 x_3 - d_3 x_2$; and similarly for other symbols, by cyclic permutation (of the indices).}
\vspace{3ex}
\end{english}
\begin{german}
Es ist nun die Bedingung aufzustellen, daß die genannten Strahlenseptupel projektiv seien.
Wir stellen die gleichwertige Bedingung auf, daß die Schnittpunkte dieser Strahlenpaare mit $\overline{a\,b}$ beziehungsweise $\overline{a'\,b'}$ projektiv seien.
Die Koordinaten dieser Schnittpunkte sind der Reihe nach:\footnote{$(d\,x)_1=d_2 x_3 - d_3 x_2$; analog die übrigen Symbole durch zyklische Vertauschung.}
\end{german}
\end{translated}
\newpage
\begin{equation*}
[1,0];\; [0,1];\; [x_1,x_2];\; [-(dx)_2,(dx)_1];\; [-(ex)_2, (ex)_1],
\end{equation*}
\begin{translated}
\begin{english}
\noindent in case of the intersections of the rays $\overline{oa}, \ldots, \overline{oe}$ 
\revTransNote{(with the line $\overline{ab}=(0,0,1)^\top$)}, 
and the roots of the equation
\end{english}
\begin{german}
\noindent für die Schnittpunkte der Strahlen $o(a\,b\,c\,d\,e)$ und die Wurzeln einer Gleichung:
\end{german}
\end{translated}
\begin{equation*}
A_{11} y^2_1 + 2 A_{12} y_1 y_2  + A_{22} y^2_2 = 0
\end{equation*}
\begin{translated}
\begin{english}
\noindent in case of the intersections of the tangent pairs 
\revTransNote{(with the line $\overline{ab}$)}, where $y_3=0$ everywhere.
\end{english}
\begin{german}
\noindent für die Schnittpunkte des Tangentenpaares, wobei überall $y_3=0$ ist.
\end{german}

\begin{english}
The corresponding values in $\epsilon'$ are denoted with prime~$'$ notation.
\end{english}
\begin{german}
Die entsprechenden Werte in $\epsilon'$ werden durch Striche bezeichnet.
\end{german}

\begin{english}
Let \revTransNote{(the entries of the dual conic of $I$, i.e., the DIAC~$I^\ast$, be 
$I^\ast \propto \footnotesize{\left(\begin{array}{ccc}
-\delta_{23} & \delta_{3} & \delta_{2}\\
\delta_{3} & -\delta_{13} & \delta_{1}\\
\delta_{2} & \delta_{1} & -\delta_{12}
\end{array}\right)}$, with entries given in terms of the IAC, $I\propto(a_{ij})$)}
\end{english}
\begin{german}
Bezeichnet man mit 
\end{german}
\end{translated}
\begin{align*}
\delta_{ik} &= a_{ii} a_{kk} - a_{ik}^2,\\
\delta_1 &= a_{11} a_{32} - a_{31} a_{12},\\
\delta_2 &= a_{22} a_{13} - a_{12} a_{23},\\
\delta_3 &= a_{33} a_{21} - a_{23} a_{31},
\end{align*}
\begin{translated}
\begin{english}
\noindent then \revTransNote{(the coefficients of the above quadratic form are)}
\end{english}
\begin{german}
% \begin{eqnarray*}
% &\delta_{ik} = a_{ii} a_{kk} - a_{ik}^2, \\
% &\delta_1 = a_{11} a_{32} - a_{31} a_{12},~\delta_2 = a_{22} a_{13} - a_{12} a_{23}, \\ &\delta_3 = a_{33} a_{21} - a_{23} a_{31},
% \end{eqnarray*}
\noindent so ergibt sich für:
\end{german}

\end{translated}
\begin{equation}
\label{eqn:OrigPaperOne}
\begin{split}
A_{11} &= \delta_{12} x^2_2 + \delta_{13} x^2_3 + 2\delta_1 x_2 x_3, \\
A_{12} &= \delta_3 x^2_3 - \delta_{12} x_1 x_2 - \delta_1 x_1 x_3 - \delta_2 x_2 x_3, \\
A_{22} &= \delta_{12} x^2_1 + \delta_{23} x^2_3 + 2\delta_2 x_1 x_3.
\end{split}
\end{equation}
\begin{translated}

\begin{english}
Since the fundamental points $(1, 0, 0)$ and $(0, 1, 0)$ correspond to one another in the projective assignment of the 7-tuple of points, this can be represented by:
\revTransNote{(Let us denote the homography between the 7-tuple of points on the lines with homogeneous coordinates $(0,0,1)^\top$ 
in each image plane ($\epsilon$ and $\epsilon'$) by the homogeneous $2\times 2$ matrix $\mH = (h_{ij})$ (representing a 1D projective transformation, of points on a single line). 
Then, $a'\propto \mH a$, $b'\propto \mH b$, i.e.,)}
\end{english}
\begin{german}
Da in der projektiven Zuordnung der Punktseptupel die Fundamentalpunkte $(1, 0, 0)$ und $(0, 1, 0)$ einander entsprechen, so läßt sich diese Projektivität darstellen durch:
\end{german}

\end{translated}
\begin{equation}
\label{eqn:OrigPaperTwo}
\begin{aligned}
\rho y_1' &= h_{11} y_1 \\
\rho y_2' &= h_{22} y_2.
\end{aligned}
\end{equation}
\begin{translated}

\begin{english}
\noindent \revTransNote{(where $h_{12}=0=h_{21}$, and $\rho$ is a non-zero scalar).} 
It follows that \revTransNote{(the equations corresponding to the intersection points associated to $c,d,e$ and $I$ are)}
\end{english}
\begin{german}
Daraus folgt:
\end{german}

\end{translated}
\begin{alignat}{8}
\lambda x_1' &= h_{11} x_1  \qquad && \mu (d' x')_2 &= h_{11} (d x)_2  \qquad && \nu (e' x')_2 &= h_{11} (e x)_2 \nonumber \\ 
\lambda x_2' &= h_{22} x_2  \qquad && \mu (d' x')_1 &= h_{22} (d x)_1  \qquad && \nu (e' x')_1 &= h_{22} (e x)_1 \label{eqn:OrigPaperThree}\\
\sigma A_{11} &= h_{11}^2 A_{11}'; \qquad && \quad \sigma A_{12} &= h_{11} h_{22} A_{12}'; \quad && \sigma A_{22} &= h_{22}^2 A_{22}' \nonumber
\end{alignat}
\begin{translated}

\begin{english}
\noindent 
\revTransNote{(where $\lambda,\mu,\nu,\sigma$ are non-zero scalars),} 
and therefore \revTransNote{(getting rid of the proportionality constants in~\eqref{eqn:OrigPaperThree})}
\end{english}
\begin{german}
\noindent und daher: 
\end{german}

\end{translated}

\begin{equation}
\label{eqn:OrigPaperFour}
\begin{aligned}
\frac{d_3 x_1 - d_1 x_3}{d_3 x_2 - d_2 x_3} \cdot \frac{x_2}{x_1} = 
\frac{d_3' x_1' - d_1' x_3'}{d_3' x_2' - d_2' x_3'} \cdot \frac{x_2'}{x_1'} \\
\frac{e_3 x_1 - e_1 x_3}{e_3 x_2 - e_2 x_3} \cdot \frac{x_2}{x_1} = 
\frac{e_3' x_1' - e_1' x_3'}{e_3' x_2' - e_2' x_3'} \cdot \frac{x_2'}{x_1'} 
\end{aligned}
\end{equation}
\addtocounter{equation}{-1}
\begin{subequations}
\begin{equation}
\label{eqn:OrigPaperFourA}
\begin{aligned}
\frac{\delta_{12} x_2^2 + \delta_{13} x_3^2 + 2 \delta_1 x_2 x_3}{\delta_3 x_3^2 - \delta_{12} x_1 x_2 - \delta_1 x_1 x_3 - \delta_2 x_2 x_3} \cdot \frac{x_1}{x_2} =
\frac{\delta_{12}' x_2'^2 + \delta_{13}' x_3'^2 + 2 \delta_1' x_2' x_3'}{\delta_3' x_3'^2 - \delta_{12}' x_1' x_2' - \delta_1' x_1' x_3' -  \delta_2' x_2' x_3'} \cdot \frac{x_1'}{x_2'}
\end{aligned}
\end{equation}
\begin{equation}
\label{eqn:OrigPaperFourB}
\begin{aligned}
\frac{\delta_{12} x_1^2 + \delta_{23} x_3^2 + 2 \delta_2 x_1 x_3}{\delta_3 x_3^2 - \delta_{12} x_1 x_2 - \delta_1 x_1 x_3 - \delta_2 x_2 x_3} \cdot \frac{x_2}{x_1} =
\frac{\delta_{12}' x_1'^2 + \delta_{23}' x_3'^2 + 2 \delta_2' x_1' x_3'}{\delta_3' x_3'^2 - \delta_{12}' x_1' x_2' - \delta_1' x_1' x_3' -  \delta_2' x_2' x_3'} \cdot \frac{x_2'}{x_1'}.
\end{aligned}
\end{equation}
\end{subequations}

\begin{translated}

\begin{english}
These four equations are satisfied by the coordinates $x_i$ and $x_i'$ of the sought points $o$ and $o'$.
We define the solutions of the ``general case'' by the requirement that none of them are zero, infinity, or undetermined.
\end{english}
\begin{german}
Dieser vier Gleichungen werden von den Koordinaten $x_i$ und $x_i'$ der gesuchten Punkte $o$ und $o'$ erfüllt.
Wir definieren die Lösungen des ``allgemeinen Falles'' durch die Forderung, daß in keiner von ihnen eine Seite null, unendlich oder unbestimmt sei.
\end{german}

\begin{english}
They are simplified by the following substitutions $\boldsymbol{\Phi}_1$ and $\boldsymbol{\Phi}_2$ 
\revTransNote{(The notation 
$x_1 : x_2 : x_3 = u_2 u_3 : u_3 u_1 : u_1 u_2$ in~\eqref{eqn:OrigPaperFive} states the proportionality of vectors typical of projective coordinates,
$(x_1,x_2,x_3)^\top\propto(u_2 u_3, u_3 u_1,u_1 u_2)^\top$.
The transformations in~\eqref{eqn:OrigPaperFive} are called reciprocal transformations, 
and are a particular type of Cremona transformations~\citep[p. 231]{Semple98book})}:
\end{english}
\begin{german}
Sie vereinfachen sich durch die folgenden Substitutionen $\boldsymbol{\Phi}_1$ und $\boldsymbol{\Phi}_2$:
\end{german}

\end{translated}
\begin{equation}
\label{eqn:OrigPaperFive}
\begin{aligned}
\boldsymbol{\Phi}_1: \quad x_1 : x_2 : x_3    &= u_2 u_3 : u_3 u_1 : u_1 u_2 \\
\boldsymbol{\Phi}_2: \quad x_1' : x_2' : x_3' &= u_2' u_3' : u_3' u_1' : u_1' u_2',
\end{aligned}
\end{equation}
\begin{translated}

\begin{english}
\noindent through which they become:
\end{english}
\begin{german}
\noindent durch die sie übergehen in:
\end{german}

\end{translated}
\begin{equation}
\label{eq:Kruppa6}
\begin{aligned}
(d_1 u_1 - d_3 u_3) (d_2'u_2' - d_3' u_3') &= (d_1' u_1' - d_3' u_3') (d_2 u_2 - d_3 u_3) \\
(e_1 u_1 - e_3 u_3) (e_2'u_2' - e_3' u_3') &= (e_1' u_1' - e_3' u_3') (e_2 u_2 - e_3 u_3) \\
\end{aligned}
\end{equation}
\addtocounter{equation}{-1}
\begin{subequations}
\begin{equation} \label{eq:Kruppa6a}
\begin{aligned}
(\delta_{12} u_3^2 + \delta_{13} u_2^2 + 2 \delta_1 u_2 u_3) \cdot 
(\delta_12' u_3'^2 + \delta_1' u_2' u_3' + \delta_2' u_1' u_3' - \delta_3' u_1' u_2') = \\
= (\delta_{12} u_3'^2 + \delta_{13}' u_2'^2 + 2 \delta_1' u_2' u_3') \cdot
(\delta_{12} u_3^2 + \delta_1 u_2 u_3 + \delta_2 u_1 u_3 - \delta_3 u_1 u_2)
\end{aligned}
\end{equation}
\begin{equation} \label{eq:Kruppa6b}
\begin{aligned}
(\delta_{12} u_3^2 + \delta_{13} u_1^2 + 2 \delta_2 u_1 u_3) \cdot 
(\delta_12' u_3'^2 + \delta_1' u_2' u_3' + \delta_2' u_1' u_3' - \delta_3' u_1' u_2') = \\
= (\delta_{12} u_3'^2 + \delta_{23}' u_1'^2 + 2 \delta_2' u_1' u_3') \cdot
(\delta_{12} u_3^2 + \delta_1 u_2 u_3 + \delta_2 u_1 u_3 - \delta_3 u_1 u_2).
\end{aligned}
\end{equation}
\end{subequations}
\begin{translated}

\begin{english}
Equation \eqref{eq:Kruppa6} was shortened by $u_1 u_2 u_3 u_1' u_2' u_3'$, 
\eqref{eq:Kruppa6a} by $u_1^2 u_2 u_3 u_1'^2 u_2' u_3'$, and 
\eqref{eq:Kruppa6b} by $u_1 u_2^2 u_3 u_1' u_2'^2 u_3'$.
\end{english}
\begin{german}
\noindent
\eqref{eq:Kruppa6} wurde durch $u_1 u_2 u_3 u_1' u_2' u_3'$, 
\eqref{eq:Kruppa6a} durch $u_1^2 u_2 u_3 u_1'^2 u_2' u_3'$, 
\eqref{eq:Kruppa6b} durch $u_1 u_2^2 u_3 u_1' u_2'^2 u_3'$ gekürzt.
\end{german}

\begin{english}
These four equations are satisfied by the coordinates $u_i$ and $u_i'$ of the points $\overline{o}$ and $\overline{o'}$ assigned to the points $o$ and $o'$ by $\boldsymbol{\Phi}_1$ and $\boldsymbol{\Phi}_2$.
To determine their number \revTransNote{(of solutions)}, we interpret these equations as follows.
The two equations~\eqref{eq:Kruppa6} represent a quadratic, birational transformation $\boldsymbol{\Sigma}$ 
\revTransNote{\citep[p. 230]{Semple98book}} between the planes $\boldsymbol{\varepsilon}$ and $\boldsymbol{\varepsilon}'$, 
which can also be written as follows:\footnote{$[d u]_1 = d_2 u_2 - d_3 u_3$, and similarly for other symbols by cyclic permutation: 
$[d u]_2 = d_3 u_3 - d_1 u_1$ and $[d u]_3 = d_1 u_1 - d_2 u_2$.}
\vspace{1ex}
\end{english}
\begin{german}
Diese vier Gleichungen werden von den Koordinaten $u_i$ und $u_i'$ der den Punkten $o$ und $o'$ durch $\boldsymbol{\Phi}_1$ und $\boldsymbol{\Phi}_2$ zugeordneten Punkte $\overline{o}$ und $\overline{o'}$ erfüllt.
Um ihre Anzahl festzustellen, interpretieren wir diese Gleichungen wie folgt.
Die zwei Gleichungen~\eqref{eq:Kruppa6} stellen eine quadratische, birationale Transformation $\boldsymbol{\Sigma}$ zwischen den Feldern $\boldsymbol{\varepsilon}$ und $\boldsymbol{\varepsilon}'$ vor, die sich auch so schreiben lässt:\footnote{$[d u]_1 = d_2 u_2 - d_3 u_3$, analog die übrigen Symbole durch zyklische Vertauschung.}
\end{german}

\end{translated}
\begin{equation} \label{eq:Kruppa7}
\begin{aligned}
u_1' : u_2' : u_3' = &\{ d_2' e_3' [d u]_2 [e u]_3 - d_3' e_2' [d u]_3 [e u]_2\} \\
: \ &\{ d_3' e_1' [d u]_3 [e u]_1 - d_1' e_3' [d u]_1 [e u]_3 \} \\
: \ &\{ d_1' e_2' [d u]_1 [e u]_2 - d_2' e_1' [d u]_2 [e u]_1 \}
\end{aligned}
\end{equation}
\begin{translated}

\begin{english}
If these conditions are substituted in~\eqref{eq:Kruppa6a} and~\eqref{eq:Kruppa6b}, we obtain the equations of two curves of 6th order $A$ and $B$, among whose 36 intersection points the points $\overline{o}$ must be found.
Similarly, two 6th order curves $A'$ and $B'$ exist in $\boldsymbol{\varepsilon}'$, with the same relevance.
By $\boldsymbol{\Sigma}$, 
$A$ and $B$ are uniquely related to $A'$ and $B'$, respectively, 
hence the points $\overline{o}$ and $\overline{o'}$ are uniquely related as well.
\end{english}
\begin{german}
Führt man diese Verhältnisse in~\eqref{eq:Kruppa6a} und~\eqref{eq:Kruppa6b} ein, so erhält man die Gleichungen von zwei Kurven 6.~Ordnung $A$ und $B$, unter deren 36~Schnittpunkten die gesuchten Punkte $\overline{o}$ vorkommen müssen.
Ebenso existieren in $\boldsymbol{\varepsilon}'$ zwei Kurven 6.~Ordnung $A'$ und $B'$ von der entsprechenden Bedeutung.
Es sind durch $\boldsymbol{\Sigma}$ $A$ auf $A'$ und $B$ auf $B'$ und daher die Punkte $\overline{o}$ auf die Punkte $\overline{o'}$ ein-eindeutig bezogen.
\end{german}

\begin{english}
Let us examine how many of the 36~intersection points of $A$ and $B$ lead to solutions to our problem.
\end{english}
\begin{german}
Es ist nun zu untersuchen, wie viele von den 36~Schnittpunkten von $A$ und $B$ zu Lösungen unseres Problems führen.
\end{german}

\begin{english}
Among these points are those which, although they are algebraic solutions of the system of equations~\eqref{eq:Kruppa6}, \eqref{eq:Kruppa6a}, \eqref{eq:Kruppa6b} (before the reduction), 
do not satisfy the conditions for the solutions of the ``general case'' with any choice of the given elements;
they are points which make both sides in the unabridged equations~\eqref{eq:Kruppa6}, \eqref{eq:Kruppa6a}, \eqref{eq:Kruppa6b} individually equal to zero.
We will now look at these points and realize that they are not solutions to our problem.
\end{english}
\begin{german}
Unter diesen befinden sich solche, die, trotzdem sie algebraische Lösungen des Gleichungssystems~\eqref{eq:Kruppa6}, \eqref{eq:Kruppa6a}, \eqref{eq:Kruppa6b} (vor der Kürzung) sind, den Bedingungen für die Lösungen des ``allgemeinen Falles'' nicht genügen bei beliebiger Wahl der gegebenen Elemente; es sind solche Punkte, die beide Seiten in den ungekürzten Gleichungen~\eqref{eq:Kruppa6}, \eqref{eq:Kruppa6a}, \eqref{eq:Kruppa6b}, einzeln zu null machen.
Wir werden nun diese Punkte aufsuchen und erkennen, daß sie keine Lösungen unseres Problems sind.
\end{german}

\begin{english}
For $u_3=0$, it follows from~\eqref{eq:Kruppa7} that
\end{english}
\begin{german}
Für $u_3=0$ folgt aus~\eqref{eq:Kruppa7}
\end{german}

\end{translated}
\begin{equation}
\begin{aligned}
u_1' : u_2' : u_3' = & u_1 [e_1 d_1 (e' d')_1 u_1 + (e_2 d_1 e_3' d_2' - e_1 d_2 e_2' d_3') u_2] \\
: \ & u_2 [e_2 d_2 (e' d')_2 u_2 + (e_2 d_1 e_1' d_3' - e_1 d_2 e_3' d_1') u_1]\\
: \ & u_1 u_2 [e_2 d_1 e_1' d_2' - e_1 d_2 e_2' d_1'].
\end{aligned}
\end{equation}
\begin{translated}

\begin{english}
Substituting these relations into~\eqref{eq:Kruppa6a} and setting $u_3 = 0$, we see that $u_1 u_2^3$ can be removed, 
while the remaining form of the second degree does not in general vanish.
It can be seen that, if~\eqref{eq:Kruppa7} is substituted in~\eqref{eq:Kruppa6a}, the terms with $u_1^6$, $u_1^5$, $u_1^4$, $u_2^6$ are absent in the equation of $A$.
From these remarks, it follows: $A$ has a triple point in $a$, and a simple point in $b$.
Correspondingly, $B$ has a simple point in $a$ and a triple point in $b$.
Assuming that $u_1 = u_2 = 0$ in~\eqref{eq:Kruppa6a} and~\eqref{eq:Kruppa7}, the equation of $A$ is satisfied.
Therefore, $A$ passes through $c$ at least once, and so does $B$.
That $c$ is only a \emph{simple point} of $A$ ($B$) can be realized as follows:
substituting $u_2 = 0$ in~\eqref{eq:Kruppa6a}, then $\delta_{12}'$ only appears in the term $\delta_{12}' \delta_2 u_1 u_3 u_3'^2$;
so if $u_1$ is removed, the remaining form does in general not identically vanish and therefore $u_1=0$ is a simple root.
Further mutual intersections cannot lie on the fundamental triangle $a,b,c$ 
since $\delta_{13}$ and $\delta_{13}'$ only appear in~\eqref{eq:Kruppa6a}, 
and $\delta_{23}$ and $\delta_{23}'$ only in~\eqref{eq:Kruppa6b}.
Thus, we have shown 7 intersections that lead to no solution.
\end{english}
\begin{german}
Setzt man diese Verhältnisse in~\eqref{eq:Kruppa6a} ein und macht $u_3 = 0$, so sieht man, daß sich $u_1 u_2^3$ herausheben läßt, während die übrigbleibende Form 2.~Grades i. a. nicht identisch verschwindet; denkt man sich weiter~\eqref{eq:Kruppa7} in~\eqref{eq:Kruppa6a} substituiert, so erkennt man, daß in der Gleichung von $A$ die Glieder mit $u_1^6$, $u_1^5$, $u_1^4$, $u_2^6$ fehlen.
Aus diesen Bemerkungen folgt: $A$ hat in $a$ einen dreifachen, in $b$ einen einfachen Punkt.
Entsprechend findet man: $B$ hat in $a$ einen einfachen, in $b$ einen dreifachen Punkt.
Setzt man in~\eqref{eq:Kruppa6a} und~\eqref{eq:Kruppa7} $u_1 = u_2 = 0$, so ist die Gleichung von $A$ befriedigt. 
$A$ geht daher wenigstens einmal durch $c$, ebenso $B$.
Dass $c$ nur ein \emph{einfacher Punkt} von $A$ ($B$) ist, erkennt man so:
Setzt man in~\eqref{eq:Kruppa6a} $u_2 = 0$, so kommt $\delta_{12}'$ nur in einem Glied $\delta_{12}' \delta_2 u_1 u_3 u_3'^2$ vor; hebt man daher $u_1$ heraus, so verschwindet die übrigbleibende Form i. a. nicht identisch und daher ist $u_1 = 0$ nur eine einfache Wurzel.
Weitere gemeinsame Schnittpunkte können auf dem Fundamentaldreiseit $a, b, c$ i. a. nicht liegen, da $\delta_{13}$ und $\delta_{13}'$ nur in~\eqref{eq:Kruppa6a}, $\delta_{23}$ und $\delta_{23}'$ nur in~\eqref{eq:Kruppa6b} vorkommen. 
Wir haben somit 7~Schnittpunkte nachgewiesen, die zu keiner Lösung führen.
\end{german}

\begin{english}
Similarly, all intersections which lie in the fundamental points of the transformation $\boldsymbol{\Sigma}$ are discarded.
According to~\eqref{eq:Kruppa7}, the fundamental lines $\overline{a'b'c'}$ correspond to conic sections whose equations are obtained by setting the square forms in the braces equal to zero.
These 3 conic sections pass through the 3 fundamental points%
\footnote{cf. e.g.,
K. Doehlemann, 
\href{https://archive.org/details/geometrischetra00doehgoog}{The quadratic and higher, birational point transformations}
in (\emph{S. Schubert XXVIII}, von Göschen, Leipzig, 1908), vol. II, p.~24. 
% \emph{Doehlemann.} %Geometrische Transformationen. II Teil: Die quadratischen und höheren, birationalen punkttransformationen (\emph{S. Schubert}), vol. II, p.~24.
}
of the transformation $\boldsymbol{\Sigma}$ in the field $\boldsymbol{\varepsilon}$
and therefore the curly brackets for these points are simultaneously zero.
Two of them are immediately recognizable from~\eqref{eq:Kruppa7}: these are the points
\vspace{1ex}
\end{english}
\begin{german}
Ebenso sind alle Schnittpunkte auszuscheiden, die in den Fundamentalpunkten der Transformation $\boldsymbol{\Sigma}$ liegen. 
Nach~\eqref{eq:Kruppa7} entsprechen den Fundamentallinien $\overline{a'b'c'}$ Kegelschnitte, deren Gleichungen man erhält, indem man die quadratischen Formen in den geschweiften Klammern gleich Null setzt. 
Diese 3~Kegelschnitte gehen durch die 3~Fundamentalpunkte%
\footnote{Vgl. etwa \emph{Doehlemann.} Geometri. Transformationen (\emph{S. Schubert}), II.~Bd., p.~24.}
der Transformation $\boldsymbol{\Sigma}$ im Felde $\boldsymbol{\varepsilon}$ und daher werden die geschweiften Klammern für diese Punkte gleichzeitig Null.
Zwei von ihnen sind aus~\eqref{eq:Kruppa7} sofort zu erkennen: es sind die Punkte
\end{german}

\end{translated}
\begin{equation*}
\overline{d} \left( \frac{1}{d_1}, \frac{1}{d_2}, \frac{1}{d_3} \right)
\end{equation*}
\begin{translated}

\begin{english}
\noindent and
\end{english}
\begin{german}
\noindent und
\end{german}

\end{translated}
\begin{equation*}
\overline{e} \left( \frac{1}{e_1}, \frac{1}{e_2}, \frac{1}{e_3} \right),
\end{equation*}
\begin{translated}

\begin{english}
\noindent that is, the points assigned to the points $d$ and $e$ by means of $\boldsymbol{\Phi}_1$.
The third (fundamental point) is the point $\overline{p}$ corresponding, through $\boldsymbol{\Phi}_1$, to the ``connected pole''%
\footnote{\emph{R. Sturm} (1869), citation above, p.~536.}
$p$ of $a b c d e$.
Again, it is evident that $d$, $e$, and $p$ cannot be solutions in general.
We now show that $\overline{d}, \overline{e}$ and $\overline{p}$ are in general double points of $A$ and $B$.
Since the 6th-order curves $A \to A'$ and $B \to B'$ are related by $\boldsymbol{\Sigma}$, 
each of them must pass through the fundamental points 6 times in all.
From~\eqref{eq:Kruppa6a} and~\eqref{eq:Kruppa7} it can be seen that $\overline{d}$ and $\overline{e}$ are double points of $A$ and $B$, hence, so is $\overline{p}$.
Therefore, 12 intersections are gathered in these three points.
\end{english}
\begin{german}
\noindent also die den Punkten $d$ und $e$ durch $\boldsymbol{\Phi}_1$ zugeordneten Punkte.
Der dritte ist der dem ``verbundenen Pol''%
\footnote{\emph{Sturm}, a. a. O., p.~536.}
$p$ von $a b c d e$ durch $\boldsymbol{\Phi}_1$ entsprechende Punkt $\overline{p}$.
Wieder ist es evident, daß i. a. $d$, $e$ und $p$ keine Lösungen sein können.
Wir zeigen nun, daß $\overline{d, \ e}$ und $\overline{p}$ i. a. gemeinsame Doppelpunkte von $A$ und $B$ sind.
Da die Kurven 6.~Ordnung $A \to A'$ und $B \to B'$ durch $\boldsymbol{\Sigma}$ aufeinander bezogen sind, muß jede von ihnen im ganzen 6mal durch die Fundamentalpunkte gehen.
Aus~\eqref{eq:Kruppa6a} und~\eqref{eq:Kruppa7} erkennt man, daß $\overline{d}$ und $\overline{e}$ Doppelpunkte von $A$ und $B$ sind, daher auch $\overline{p}$.
In diesen drei Punkten sind daher 12~Schnittpunkte vereinigt.
\end{german}

\begin{english}
We now examine whether the equations~\eqref{eq:Kruppa6a} and~\eqref{eq:Kruppa6b} also allow to discard intersection points.
These equations are of the form
\vspace{1ex}
\end{english}
\begin{german}
Wir untersuchen nun,, ob auch die Gleichungen~\eqref{eq:Kruppa6a} und~\eqref{eq:Kruppa6b} zur Ausscheidung von Schnittpunkten Veranlassung geben.
Diese Gleichungen sind von der Form
\end{german}

\end{translated}
\begin{equation} \label{eq:Kruppa9}
M C' = M' C \quad \text{and/und} \quad N C' = N' C.
\end{equation}
\begin{translated}

\begin{english}
On the conic section $C$ there are 8 points, which are assigned to points on $C'$ by means of $\boldsymbol{\Sigma}$.
Equations~\eqref{eq:Kruppa6}, \eqref{eq:Kruppa6a} und~\eqref{eq:Kruppa6b} are simultaneously satisfied by the coordinates of these 8 point pairs.
$A$ and $B$ therefore intersect $C$ in 8 points.
Now, however, $C'$ passes through $a'$ and $b'$, which are assigned to $a$ and $b$ by $\boldsymbol{\Sigma}$, according to~\eqref{eq:Kruppa7}.
However, since we have already discarded $a$ and $b$, we only get to know 6 new intersections.
Further consideration of~\eqref{eq:Kruppa9} shows that the points of $A$ which make the two sides of the equation of $A$ equal to zero, individually, do not in general lie on $B$, and vice versa.
Therefore, the inspection is complete.
\end{english}
\begin{german}
Auf dem Kegelschnitt $C$ liegen 8~Punkte, denen durch $\boldsymbol{\Sigma}$ Punkte zugeordnet sind, die auf $C'$ liegen. 
Für die Koordinaten dieser 8~Punktpaare sind die Gleichungen~\eqref{eq:Kruppa6}, \eqref{eq:Kruppa6a} und~\eqref{eq:Kruppa6b} gleichzeitig erfüllt.
$A$ und $B$ schneiden sich daher in 8~Punkten auf $C$.
Nun geht aber $C'$ durch $a'$ und $b'$, denen durch $\boldsymbol{\Sigma}$ nach~\eqref{eq:Kruppa7} $a$ und $b$ zugeordnet sind.
Da wir  aber $a$ und $b$ bereits entfernt haben, lernen wir daher bloß 6 neue auszuscheidende Schnittpunkte kennen.
Die weitere Betrachtung von~\eqref{eq:Kruppa9} zeigt, daß die Punkte von $A$, die die beiden Seiten der Gleichung von $A$ einzeln zu Null machen, i. a. nicht auf $B$ liegen und umgekehrt. 
Daher ist die Untersuchung abgeschlossen.
\end{german}

\begin{english}
In total, we excluded 25 points.
Therefore, the problem has in general 11 solutions, regardless of whether they are real or complex.
\end{english}
\begin{german}
Im ganzen haben wir 25~Punkte ausgeschlossen.
Daher hat das Problem i. a. ohne Rücksicht auf die Realität 11~Lösungen.
\end{german}

\begin{english}
We now show that for every solution $o, o'$ there are 2 possible orientations of the perspective views.
We extract one of the two tangents $T$ to $I$ from the epipolar line pencil \revTransNote{(defined by)} $o$.
Through the projective relation between the epipolar line pencils \revTransNote{(the epipolar line homography)} it uniquely corresponds to a tangent $T'$ to $I'$.
We now bring the two systems $(\boldsymbol{\varepsilon} o_1)$ and $(\boldsymbol{\varepsilon}' o_2)$ into a position such that the lines $\overline{o_1 o}$ and $\overline{o_2 o'}$ coincide and that the planes spanned by $[T,o_1]$ and $[T',o_2]$ also coincide.
This condition, however, determines the relative position of the systems $(\boldsymbol{\varepsilon} o_1)$ and $(\boldsymbol{\varepsilon}' o_2)$ only up to \emph{screw transformations} about the axis $\overline{o_1 o_2}$.
We can therefore establish the new condition that planes (containing points) $[o_1 o_2 a]$ and $[o_1 o_2 a']$ also coincide.
Then $a$ and $a'$ are actually the central projections of a space point $a^r$, 
and it is now necessary to prove that the other pairs of image points $b c d e \to b' c 'd' e '$ also correspond to space points $b^r c^r d^r e^r$.
The two epipolar line pencils already have such a position that $\overline{o a}$ and $\overline{o' a'}$ also meet $T$ and $T'$ on the intersection line of $\boldsymbol{\varepsilon}$ and~$\boldsymbol{\varepsilon}'$.
But since the minimum cones with vertices $o_1$ and $o_2$ have two common tangential planes $\boldsymbol{\tau}_1$ and $\boldsymbol{\tau}_2$, the other tangents from $o$ and $o'$ to $I$ and $I'$ must also intersect each other on this line.
According to the fundamentals of the projective geometry, the two epipolar line pencils therefore are projectively related and therefore the perspective views are orientated.
\end{english}
\begin{german}
Wir zeigen nun, daß zu jeder Lösung $o, o'$ 2~Orientierungsmöglichkeiten der Perspektiven gehören.
Wir greifen aus dem Kernstrahlbüschel $o$ eine der beiden Tangenten $T$ an $I$ heraus.
In der Projektivität der Kernstrahlbüschel entspricht ihr eindeutig eine Tangente $T'$ an $I'$.
Wir bringen nun die beiden Systeme $(\boldsymbol{\varepsilon} o_1)$ und $(\boldsymbol{\varepsilon}' o_2)$ in solche Lage, daß die Geraden $\overline{o_1 o}$ und $\overline{o_2 o'}$ zusammenfallen und daß sich die Minimalebenen $[T o_1]$ und $[T' o_2]$ decken.
Durch diese Bedingung ist aber die relative Lage der Systeme $(\boldsymbol{\varepsilon} o_1)$ und $(\boldsymbol{\varepsilon}' o_2)$ nur bis auf die \emph{Schraubungen} um die Achse $\overline{o_1 o_2}$ bestimmt.
Wir können daher die neue Bedingung einführen, daß auch die Ebenen $[o_1 o_2 a]$ und $[o_1 o_2 a']$ zusammenfallen.
Dann sind $a$ und $a'$ tatsächlich die Zentralrisse ein Raumpunktes $a^r$, und es ist nun zu beweisen, daß auch die anderen Bildpaare $b c d e \to b' c' d' e'$ zu Raumpunkten $b^r c^r d^r e^r$ gehören.
Die beiden Kernstrahlbüschel haben bereits solche Lage, daß sich $\overline{o a}$ und $\overline{o' a'}$, ferner $T$ und $T'$ auf der Schnittlinie von $\boldsymbol{\varepsilon}$ und $\boldsymbol{\varepsilon}'$ treffen.
Da aber die Minimalkegel mit den Spitzen $o_1$ und $o_2$ zwei gemeinsame Tangentialebenen $\boldsymbol{\tau}_1$ und $\boldsymbol{\tau}_2$ haben, müssen auch die anderen Tangenten aus $o$ und $o'$ an $I$, beziehungsweise $I'$ einander auf dieser Geraden schneiden.
Nach dem Fundamentalsatz der projektiven Geometrie liegen daher die beiden Kernstrahlbüschel perspektiv und daher sind die Perspektiven in orientierter Lage.
\end{german}

\begin{english}
The question yet to be answered is to how many possible orientations can a pair of epipoles belong.
First, there is the freedom to choose the distance $\overline{o_1 o_2}$ arbitrarily.
As a result, it is not possible to determine the \emph{true size} but only the \emph{shape} of the pentagon.
If the distance $\overline{o_1 o_2}$ is given and the orientation is carried out in a way, 
then a second orientation is obtained by rotating the system $(\boldsymbol{\varepsilon}' o_2)$ around $\overline{o_1 o_2}$ by $180^\circ$, 
since this rotation maps the epipolar planes $\overline{o_1 o_2} (a b c d e)$ onto themselves.%
\footnote{If only the principal points and the distance are entered into the concept of ``inner orientation'', 
without specifying on which side of the image plane (Euclidean plane) the center lies, 
then the reflections at the points of $\overline{o_1 o_2}$ and the normal planes of $\overline{o_1 o_2}$ are added to the above-mentioned screw transformations around $\overline{o_1 o_2}$.
After disposing of the distance $\overline{o_1 o_2}$, 
there are 8 possible orientations consistent with a pair of epipoles.}
\end{english}
\begin{german}
Es ist noch die Frage zu beantworten, wie viel Orientierungsmöglichkeiten zu einem Kernpunktpaar gehören.
Zunächst besteht die Freiheit, die Entfernung $\overline{o_1 o_2}$ beliebig zu wählen.
Dies hat zur Folge, daß sich nicht die \emph{wahre Grösse} sondern nur die \emph{Gestalt} des Fünfecks ermitteln läßt.
Ist die Entfernung $\overline{o_1 o_2}$ gewählt und die Orientierung auf \emph{eine} Weise durchgeführt, so erhält man eine zweite Orientierung, indem man das System $(\boldsymbol{\varepsilon}' o_2)$ um $\overline{o_1 o_2}$ durch $180^\circ$ umwendet, denn  bei dieser Umwendung gehen die projizierenden Ebenen $\overline{o_1 o_2} (a b c d e)$ in sich über.%
\footnote{Läßt man in den Begriff ``innere Orientierung'' bloß Hauptpunkte und Distanz eingehen, ohne anzugeben, auf welcher Seite der Bildebene (euklidische Ebene) das Zentrum liegen soll, so kommen zu den oben erwähnten Schraubungen um $\overline{o_1 o_2}$ noch die Spiegelungen an den Punkten von $\overline{o_1 o_2}$ und den Normalebenen von $\overline{o_1 o_2}$ hinzu.
Nach Verfügung über den Abstand $\overline{o_1 o_2}$ gehören dann zu einem Kernpunktpaar 8~Orientierungsmöglichkeiten.}
\end{german}

%%%%%%%%%%%%%%%%%%%%%%%%%%%%%%%%%%%%%%%%%%%%%%%%%%%%%%%%%%%

\begin{english}
\section*{Proof of \nth{2} Theorem}
The correctness of the second theorem follows easily from the auxiliary theorem~$(b)$ p.~\pageref{thm:EpipolarLineHomography}.
According to the projectivity problem for 7 point pairs, the epipoles $o\,o'$ can be determined in three ways.
If the calibration of $\epsilon$ is given, the image $I$ of the Absolute Conic $I^r$ can be regarded as known.
The center point $I'$ of $I^r$ on $\epsilon'$, which is to be determined first, must, according to $(b)$, touch the pair of lines that correspond to the tangents of $I$ from $o$ in the projectivity of the pencils of epipolar lines.
Therefore, if the distance of $o_2$ is given, $o_2$ can be determined in four ways.\footnote{\label{foot:thm2}Assuming that the center is on the side of the (Euclidean) image plane that carries the perspective; otherwise there are twice as many solutions.}
If the epipole pair is real, then out of the 4 solutions, two are real and two are complex conjugate.
For the orientation problem stated in Theorem~2, there are thus generally 24 solutions.
\end{english}
\begin{german}
\section*{Zu 2.}
Auch die Richtigkeit des zweiten Satzes folgt leicht aus dem Hilfssatz~$(b)$ p.~\pageref{thm:EpipolarLineHomography}.
Nach dem Problem der Projektivität für 7~Punktpaare läßt sich das Kernpunktpaar $o\,o'$ auf drei Arten bestimmen.
Ist die innere Orientierung von $\epsilon$ gegeben, so ist der Zentralriß~$I$ des absoluten Kegelschnittes $I^r$ als bekannt anzusehen.
Der erst zu ermittelnde Zentralriß~$I'$ von $I^r$ auf $\epsilon'$ muß nach $(b)$ das Geradenpaar berühren, das in der Projektivität der Kernstrahlbüschel dem Tangentenpaar durch $o$ an $I$ entspricht.
Ist daher noch die Distanz von $o_2$ gegeben, so ist $o_2$ auf vier Arten bestimmbar.\footnote{Unter der Voraussetzung, daß sich das Zentrum auf der Seite der (euklidischen) Bildebene befindet, die die Perspektive trägt; sonst doppelt soviel Lösungen.}
Ist das Kernpunktpaar reell, so sind von den 4~Lösungen 2~reell, 2~konjugiert imaginär.
Für das in Satz~2 enthaltene Orientierungsproblem ergeben sich somit i. a. 24~Lösungen.
\end{german}

\begin{english}
Finally, it should be pointed out that the feasibility of the two orientation problems discussed here is also suggested by the degrees-of-freedom count stated in my work on p.~1 (note).
This count also indicates the feasibility of the following problem:
to solve for the orientation given the projections of 6 space points, the internal calibration of one camera and the principal point of the other.
It can be seen that this problem is an intermediate one between the two treated here, but I have not succeeded in solving it.
\end{english}
\begin{german}
Zum Schlusse sei darauf hingewiesen, daß die Möglichkeit der beiden hier behandelten Orientierungsprobleme auch durch die in meiner auf p.~1 (Anmerkung) angeführten Arbeit erklärte Konstantenzählung nahegelegt wird.
Diese Konstantenzählung deutet aber auch auf die Möglichkeit des folgenden Problemes hin: 
Gegeben sind zwei Perspektiven von 6~Raumpunkten, die innere Orientierung der einen und der \emph{perspektive Hauptpunkt} der anderen. 
Man sieht, daß diese Problem zwischen den beiden von uns behandelten steht, doch ist mir seine Lösung nicht gelungen.
\end{german}

\begin{english}
The count of the degrees of freedom also shows that the assumption of two perspective views with an internal orientation of 4 points is generally not sufficient to determine a spatial quadrilateral. 
It is well known, however, that this problem is solvable when one knows that the 4 space points lie on a plane.
\end{english}
\begin{german}
Die Konstantenzählung ergibt ferner, daß die Annahme von 2~Perspektiven mit innerer Orientierung von 4~Punkten im allgemeinen nicht genügt, um ein räumliches Viereck zu bestimmen.
Es ist aber bekannt, daß dieses Problem bestimmt ist, wenn man weiß, daß die 4~Raumpunkte in einer Ebene liegen.
\end{german}

\begin{english}
With regard to the ambiguity of the orientation problems, the following is worth noting.
If an orientation problem is based on actual perspective images --- without arbitrary assumptions ---
of a spatially extended object,
which reveal more elements than is necessary to determine the problem, 
then the new task consists of finding the solutions by which the other elements of the picture can also be reconciled;
i.e., in general, there is then only one possible orientation.
\end{english}
\begin{german}
Bezüglich der Mehrdeutigkeit der Orientierungsprobleme ist noch folgendes zu bemerken.
Gründet man ein Orientierungsproblem auf tatsächliche Perspektiven --- nicht willkürliche Annahmen --- eines räumlich ausgedehnten Objektes, die mehr Elemente erkennen lassen, als zur Bestimmtheit des Problems erforderlich ist, so entsteht die neue Aufgabe, diejenigen Lösungen zu finden, durch die sich auch die übrigen Bildelemente in Einklang bringen lassen; i. a. gibt es dann nur eine Orientierungsmöglichkeit.
\end{german}

\end{translated}

% Kruppa's paper (original)
\pagebreak
\pagenumbering{gobble} % remove page numbers
\includepdf[pages=-,pagecommand={},landscape=true,noautoscale=false,scale=1.05,offset=7 0]{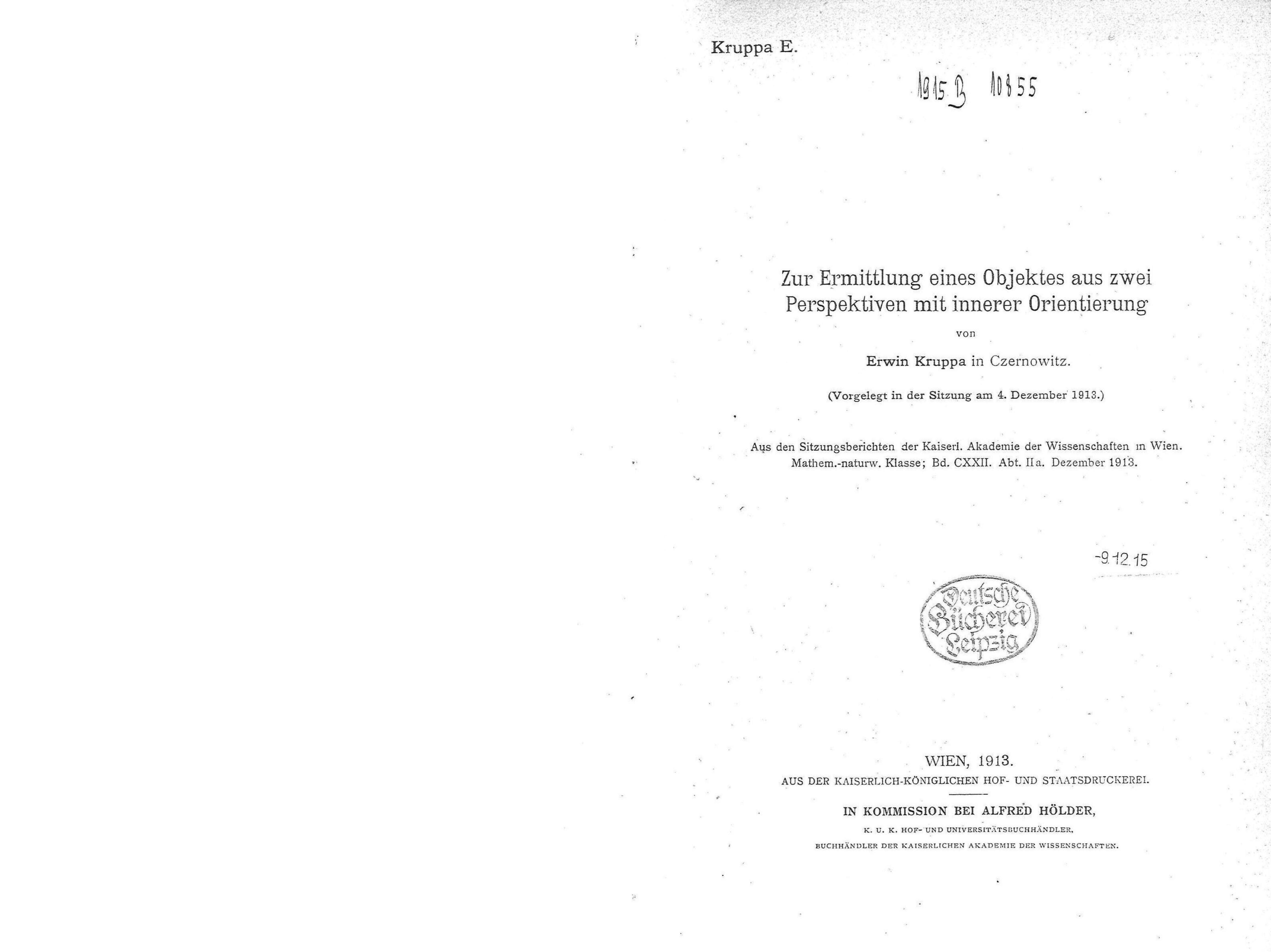}

\end{document}